\documentclass{article}
\usepackage{iclr2027_conference,times}

\usepackage{amsmath,amsfonts,bm}

\def\eqref#1{equation~\ref{#1}}

\def\1{\bm{1}}

\DeclareMathAlphabet{\mathsfit}{\encodingdefault}{\sfdefault}{m}{sl}
\SetMathAlphabet{\mathsfit}{bold}{\encodingdefault}{\sfdefault}{bx}{n}

\usepackage{amssymb}
\usepackage{placeins}
\usepackage{needspace}

\usepackage{hyperref}
\hypersetup{hidelinks}
\usepackage{url}
\usepackage{xcolor}
\usepackage{colortbl}
\definecolor{kvhl}{HTML}{E7F1FB}
\usepackage{graphicx}
\usepackage{booktabs}
\usepackage{algorithm}
\usepackage{algpseudocode}
\usepackage{enumitem}

\iclrfinalcopy

\title{KVFetch: Temporal Prefetching for the Missing Half\\of KV Cache Compression}

\author{Linfeng Dong}

\begin{document}

\maketitle
\lhead{Under review as a conference paper at ICLR 2027}

\begin{abstract}
As context windows scale to tens or hundreds of thousands of tokens,
KV cache compression has become essential for efficient LLM inference.
Existing methods fall into three families: score-based eviction, summary
compensation, and offload-and-recall. Yet all three decide what to keep
or recall by content relevance to the current query. We show this shared
design is structurally incomplete. A cache supports two access modes:
associative lookup by content and sequential traversal by position;
current compressors implement only the first. The gap matters in
practice: retrieval-augmented generation, code completion, and
structured-data extraction all require the model to reproduce
identifiers, field values, or code tokens verbatim from the context.
Under compression, content-based eviction retains the head of such a
sequence but discards its continuation, causing verbatim copying to break
irreversibly midway, a failure we call sequential forgetting. This
failure resists better scoring, larger budgets, summary compensation, and
dynamic re-scoring; it is the dominant source of remaining quality loss
under compression. We propose KVFetch, a training-free, drop-in framework
that opens a temporal recall channel for any score-based compressor. It
demotes evicted candidates to a quantized cold tier, detects active
copying through a monotone read pointer, and prefetches positional
successors into fixed-size hot-tier slots without increasing attention
cost. On RULER-16K under an iso-budget control, KVFetch recovers verbatim
copying from 0.8 to 78.4 and raises the 13-task average by \textbf{+8.4},
with gains concentrating on tasks that require sequential access. On
LongBench, where no task requires sequential access, the channel remains
dormant and imposes no cost.
\end{abstract}

\section{Introduction}
\label{sec:intro}

The rapid growth of context windows has made KV cache compression a prerequisite for
practical long-context LLM inference. Three families of methods have emerged to manage
the resulting memory cost: score-based eviction retains the highest-scoring entries and
discards the rest \citep{h2o,snapkv,obcache}, summary compensation approximates the
aggregate contribution of discarded entries \citep{indexmem,momentkv}, and
offload-and-recall retrieves entries beyond the active cache on demand
\citep{infllm,quest}. Despite their diversity, all three families share a single
addressing primitive: they decide what to keep or recall by measuring content relevance
to the current query.

We show that content addressing alone is structurally incomplete. A cache supports two
kinds of access: associative lookup by content and sequential traversal along the
position stream. Every existing compressor implements the first; none provides the
second. This incompleteness is not easily resolved by better attention scores, larger
retention budgets, or lossy summary compensation. While better scoring could in principle
change which positions are retained, our retention sweep shows that existing methods do
not preferentially preserve mid-span successors even at a 40\% budget, because these
tokens are semantically indistinguishable from other low-scoring positions. Crucially,
the sequential demand is generated by the decoding trajectory itself, making it
fundamentally difficult for any prefill-time scoring function to anticipate.

\begin{figure}[t]
\centering
\begin{minipage}[c]{0.44\linewidth}
\includegraphics[width=\linewidth]{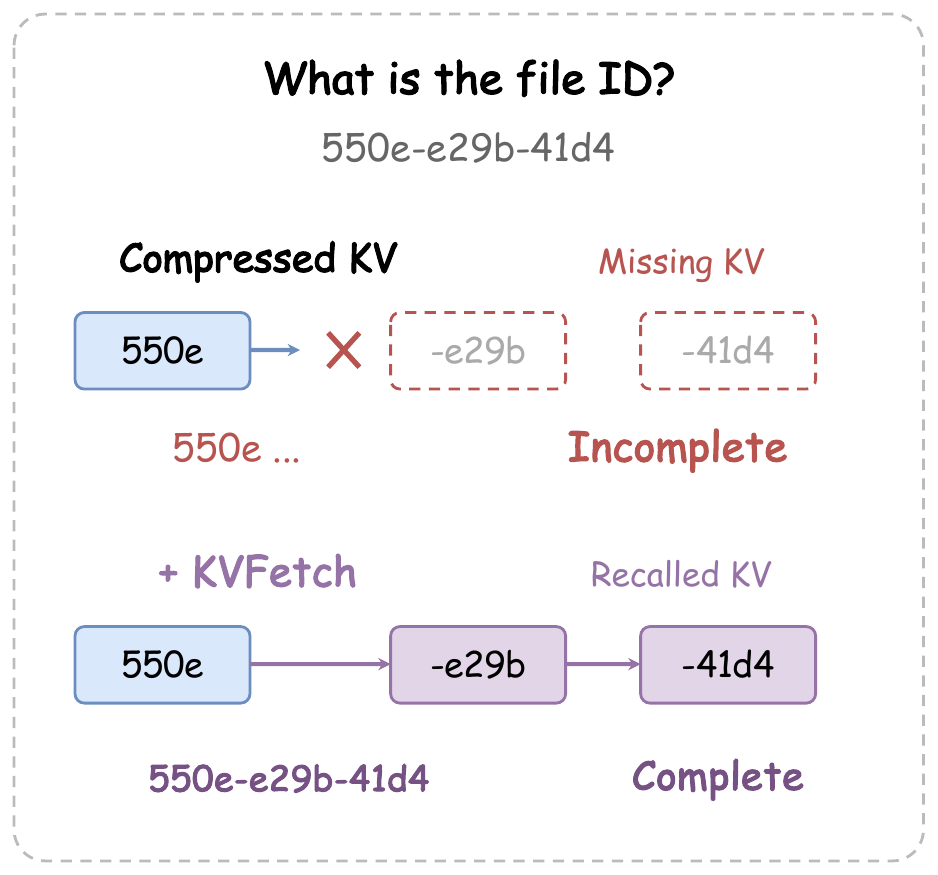}
\end{minipage}\hfill
\begin{minipage}[c]{0.43\linewidth}
\caption{Sequential forgetting and recovery with KVFetch (schematic).
\textbf{Top:} a compressed cache retains the target prefix but lacks the KV entries
for its continuation, interrupting verbatim copying.
\textbf{Bottom:} KVFetch recalls admitted successors and enables the copy to continue.
Recovery requires a visible prefix and cold-tier coverage.}
\label{fig:teaser}
\end{minipage}
\end{figure}

Figure~\ref{fig:teaser} illustrates the consequence. When the model must copy a UUID
token by token, content-based eviction retains the span's head tokens, which carry the
query-relevant semantics, but discards mid-span successors whose standalone scores are
uniformly low. The model locates the target and begins copying correctly, then breaks
irreversibly once it reaches evicted positions. We call this failure mode
\emph{sequential forgetting}. Unlike diffuse quality loss, which responds to score
refinement or larger budgets, sequential forgetting is abrupt and resists the four
natural improvement axes we examine (better scoring, larger budgets, summary
compensation, and dynamic re-scoring). In our evaluation, tasks exhibiting this pattern
account for 86\% of the strongest baseline's residual gap to uncompressed performance.
This resistance is structural: no score refinement can recover a physically deleted
position; larger budgets do not preferentially retain mid-span successors (s3 rises from
0.8 to only 3.2 at 40\%); summary methods approximate aggregate attention but not
bit-exact identity; and dynamic re-scoring still evaluates content relevance (TOVA's
s3 = 0.0). We detail this analysis in \S\ref{sec:exp-diagnosis}.

We propose that KV cache compression should implement both addressing modes, mirroring
the dual-channel design that has been standard in hardware caches for decades. KVFetch is
a concrete instantiation: a training-free, drop-in framework that opens the temporal
channel for any score-based compressor without modifying its scores, activating whenever
the model's attention trajectory exhibits sequential advance and remaining dormant
otherwise. It implements the channel through three components: \emph{admission} demotes
low-predictability literal spans (e.g., hex digits, rare identifiers) to a quantized cold
tier rather than permanently deleting them; a monotone \emph{read pointer} detects active
verbatim copying; and \emph{prefetch-and-swap} recalls positional successors into
fixed-size hot-tier slots, keeping attention cost constant.

Under an iso-budget control, KVFetch raises the RULER-16K 13-task average by +8.4 (9 of
13 tasks improve, no material regression), recovers verbatim copying from 0.8 to 78.4,
and shows no degradation on tasks without sequential access (LongBench +0.33). The gains
are consistent across three model families (Qwen2.5-7B, Qwen3-4B, Llama-3.1-8B).

\paragraph{Contributions.}
\begin{enumerate}[leftmargin=*]
\item We identify a missing addressing mode in KV cache compression. All existing
compressors implement content addressing; none provides temporal addressing. We formalize
the resulting failure as \emph{sequential forgetting} and show it accounts for 86\% of the
aggregate per-task residual gap between the strongest baseline and full KV.
\item We introduce KVFetch, a training-free temporal recall framework that adds a
position-indexed cold tier and fixed-capacity hot-slot prefetching to any score-based
compressor, without modifying its scoring logic.
\item We evaluate on long-context benchmarks (RULER, LongBench) with budget and length
scaling, iso-budget controls, and component ablations demonstrating that each of the three
components is necessary.
\end{enumerate}
\section{Related Work}
\label{sec:related}

We organize existing work by \emph{addressing primitive}, the mechanism that decides what to
keep or recall, since this primitive determines what each family can and cannot recover.

\textbf{Score-based eviction.}
The dominant family assigns importance scores and retains the top-scoring fraction. Methods
differ in when and how scores are computed: H2O~\citep{h2o} accumulates prefill attention,
SnapKV~\citep{snapkv} pools an observation window, and TOVA~\citep{tova} updates selection
during decoding. A second wave of refinements improves the selection process along various
axes: OBCache~\citep{obcache} applies perturbation-based pruning, AdaKV~\citep{adakv} and
PyramidKV~\citep{pyramidkv} allocate budgets adaptively across heads or layers,
CapKV~\citep{capkv} optimizes an information-capacity objective, CompressKV~\citep{compresskv}
selects through semantic retrieval heads, and FASA~\citep{fasa} estimates query-dependent
importance from rotary-frequency components. Despite this diversity, every method evaluates
\emph{content relevance}: how much a given position matters to the current query. This
primitive is effective for associative lookup, but improving a score cannot recover a
physically deleted position; mid-span successors of a literal span carry uniformly low scores
and are not preserved even at 40\% retention.

SnapKV's neighborhood preservation is the closest precursor to our concern: it extends a
fixed-size window around selected attention peaks to avoid losing surrounding context, such as
partial phone numbers. However, this window is a static, prefill-time decision with a fixed
radius that does not adapt to the copy trajectory emerging during decoding. For multi-token
literal spans, the window typically does not extend far enough to cover the full continuation.
KVFetch addresses the same failure class with a dynamic, decoding-time mechanism that detects
ongoing copying and recalls successors on demand.

\textbf{Summary compensation.}
IndexMem~\citep{indexmem} combines learned eviction with latent memory, MomentKV~\citep{momentkv}
uses moment statistics, and RESA~\citep{resa} approximates residual attention. These methods
recover the \emph{distributional effect} of discarded tokens on the attention output, but not
the \emph{bit-exact token identity} required for verbatim copying, which is the critical
distinction for sequential continuation.

\textbf{Offload-and-recall.}
InfLLM~\citep{infllm} retrieves blocks from external context memory, ArkVale~\citep{arkvale}
makes eviction recallable, Quest~\citep{quest} estimates page relevance from query vectors and
key extrema, and NOSA~\citep{nosa} jointly designs sparse attention with bounded CPU--GPU
transfers. Their selection criterion remains content-based: they retrieve the entries most
similar to the current query. Content-based recall can co-load positional successors
incidentally (Quest's page granularity does so, raising s3 from 0.8 to 8.8), but it does not
systematically follow the copy trajectory and remains far from the full-KV ceiling (99.2).

\textbf{Predicting future demand.}
SparDA~\citep{sparda}, FlashMemory~\citep{flashmemorylsa}, and KVP~\citep{kvp} train
predictors for future KV demand, recognizing, as we do, that demand is not fully determined at
prefill time. They target broad demand patterns with learned models; KVFetch targets a specific
pattern, sequential continuation, with a training-free signal. The two approaches are
complementary.

\textbf{KV cache quantization.}
KIVI~\citep{kivi}, KVQuant~\citep{kvquant}, and GEAR~\citep{gear} reduce bytes per retained
entry via low-precision representations. KVFetch uses 4-bit affine quantization for its cold
tier as an implementation choice, not as a quantization contribution.

Across all four axes, the addressing primitive has been content relevance. No existing method
addresses positions by their \emph{temporal relation} to the model's current read trajectory.
KVFetch provides this second channel.

\renewcommand{\eqref}[1]{(\ref{#1})}

\section{Method: KVFetch}
\label{sec:method}
\label{sec:observation}

KVFetch adds a temporal recall channel to any score-based compressor. The distinction
between content addressing and temporal addressing mirrors a classical dichotomy in
hardware cache design: associative caches index by content tag, while sequential
prefetchers predict demand from the position stream. Modern CPU caches have long combined
both modes; KV cache compressors, however, have implemented only the first. KVFetch
provides the second.

\subsection{Preliminaries and Notation}
\label{sec:method-prelim}

A score-based compressor assigns an importance score $s_p$ to each position $p$ after
prefill and retains only the top-$\lfloor(1-\rho)L\rfloor$ positions:
\[
\mathcal{H} = \operatorname{top}\text{-}\lfloor(1-\rho)L\rfloor\!\left(\{s_p\}_{p=0}^{L-1}\right)
\]
where $L$ is the prompt length and $\rho$ the eviction rate. Decoding then attends
exclusively over $\mathcal{H}$, whose membership remains fixed thereafter. We write $a_t$ for
the attention distribution at decoding step $t$, $\mathcal{H}_t$ for the hot tier, and
$\mathcal{C}$ for the cold tier introduced by KVFetch. Batch, layer, and head indices
are omitted except where semantics differ.

\subsection{Sequential Forgetting}
\label{sec:method-forgetting}

\paragraph{Definition 1 (Sequential forgetting).}
Let $\pi = (p_1, \ldots, p_m)$ be a contiguous span that must be reproduced verbatim. A
compressed cache $\mathcal{H}$ exhibits \emph{sequential forgetting} on $\pi$ if there
exists $j < m$ such that $p_1, \ldots, p_j \in \mathcal{H}$ but $p_{j+1} \notin
\mathcal{H} \cup \operatorname{dom}(\mathcal{C})$. The model generates the first $j$
tokens correctly, then fails irreversibly at position $p_{j+1}$.

Sequential forgetting differs from diffuse forgetting in two ways: (i)~it is abrupt,
meaning an otherwise correct copy breaks sharply at a single position; and (ii)~it resists
the four natural improvement axes (better scoring, larger budgets, summary compensation,
and dynamic re-scoring). Tasks exhibiting this pattern account for 86\% of the aggregate
per-task gap between the strongest baseline and full KV.

\subsection{Overview and System State}
\label{sec:method-overview}

\begin{figure*}[t]
\centering
\includegraphics[width=\textwidth]{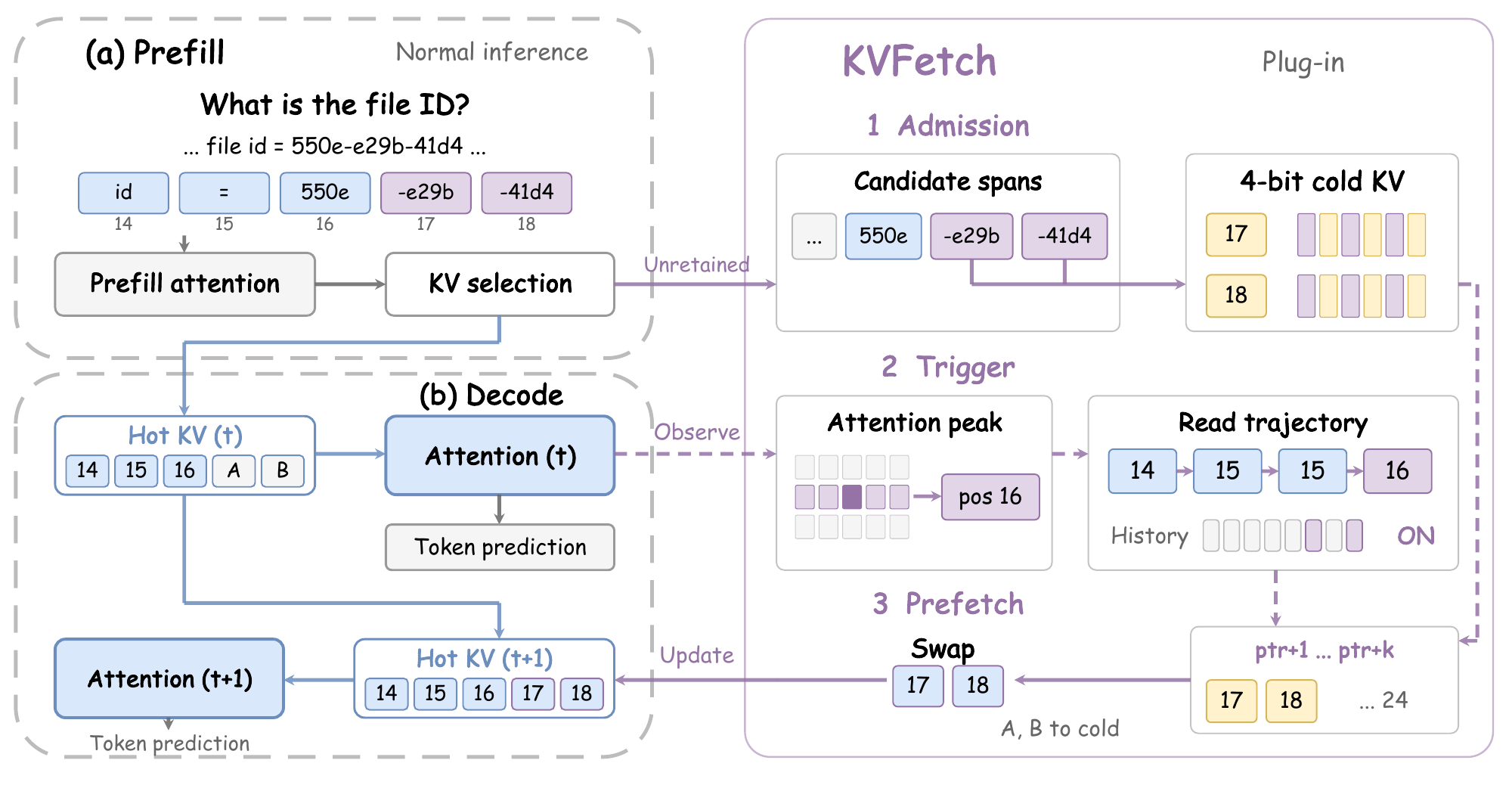}
\caption{Overview of KVFetch as a training-free plug-in.
\textbf{Left:} ordinary \textbf{(a)} prefill KV selection and \textbf{(b)} decoding.
\textbf{Right:} \textbf{(1)} unretained candidate spans enter a position-indexed 4-bit
cold tier; \textbf{(2)} attention peaks update a stall-tolerant read pointer and
trigger; \textbf{(3)} positional successors are fetched and swapped into the hot tier.
Swaps preserve the number of prompt hot slots; cold storage is additional.
Attention cells and token chunks are schematic; two fetched entries illustrate the
update (default $k=8$).}
\label{fig:overview}
\end{figure*}

KVFetch extends the baseline through three components: admission (\S3.4), triggering
(\S3.5), and prefetching (\S3.6), each described below. Figure~\ref{fig:overview}
illustrates the pipeline. The system augments the baseline state to
$\mathcal{S}_t = \left(\mathcal{H}_t,\ \mathcal{C},\ \mathcal{M}_{\mathrm{pos}},\ \mathrm{ptr}_t,\ \mathbf{h}_t\right)$,
where $\mathcal{C}$ is a 4-bit cold tier (capped at 5\% of $L$) with an $O(1)$ position
index, $\mathcal{M}_{\mathrm{pos}}$ maps hot slots to original positions, and
$\mathrm{ptr}_t$ is the read pointer. The state is updated by
$\mathcal{S}_{t+1} = \Phi(\mathcal{S}_t, \mathbf{a}_t)$ using only the attention already
computed: no extra forward passes, no training. Without sequential access, all added
components remain idle and the system reduces to the unmodified baseline.

\subsection{Admission: Constructing the Cold Tier}
\label{sec:method-admission}

Tokens lacking standalone semantics, such as hexadecimal digits, base64 fragments, and rare
identifiers, receive uniformly low importance scores at prefill because the verbatim demand
that will eventually consume them does not yet exist. Permanently deleting these positions
makes sequential forgetting inevitable.

KVFetch demotes such tokens into a 4-bit quantized cold tier indexed by original position.
The admission signal is unigram-frequency rank: tokens in the bottom-$\tau$ percentile form
candidate runs when they occur consecutively, extended by $\delta$ positions on each end for
tokenizer-boundary tolerance. This signal is format-agnostic: hex strings, JSON keys, code
identifiers, and rare nouns are all captured without format-specific rules.

The cold-tier budget is computed on demand and capped at $\gamma_{\mathrm{cold}} = 5\%$ of
$L$. Tasks without candidate spans receive no cold tier, and unused budget is partially
refunded to the hot tier, eliminating any regression on non-copying tasks. The cold tier uses
4-bit per-channel affine quantization; 4-bit matches 8-bit precision in our experiments,
while 2-bit degrades substantially (\S\ref{app:sensitivity}).

\subsection{Triggering: Detecting Active Copying}
\label{sec:method-trigger}

When the model copies a span verbatim, its attention peak advances monotonically in the
original position coordinate. We build the trigger from this behavioral signature.

At each decoding step $t$, we extract the attention peak over prompt tokens and recover
its original position:
\begin{equation}
p_t^{*} = \mathcal{M}_{\mathrm{pos}}\Big[\arg\max_{j\in\mathcal{V}} \mathbf{a}_t[j]\Big]
\label{eq:peak}
\end{equation}
Each layer maintains its own pointer, observed through its most confident KV head. Let
$\Delta_t = p_t^{*} - \mathrm{ptr}$. The pointer handles three cases:
\begin{equation}
\mathrm{ptr} \leftarrow \begin{cases}
p_t^{*}, & 0 < \Delta_t \leq \Delta_{\max} \quad (\text{advance})\\
\mathrm{ptr}, & -\Delta_{\max} \leq \Delta_t \leq 0 \quad (\text{stall})\\
p_t^{*}, & |\Delta_t| > \Delta_{\max} \quad (\text{jump; reset})
\end{cases}
\label{eq:trigger}
\end{equation}
Copy mode activates when at least $n_{\mathrm{adv}} = 2$ advances occur within a window of
$W = 8$ steps:
\begin{equation}
\mathrm{Adv}_t = \mathbf{1}\big[0 < \Delta_t \leq \Delta_{\max}\big],
\qquad
\mathrm{Copy}_t = \mathbf{1}\Big[\textstyle\sum_{i=0}^{W-1}\mathrm{Adv}_{t-i} \geq n_{\mathrm{adv}}\Big]
\label{eq:copy}
\end{equation}
with $\Delta_{\max} = 4$. The trigger activates on approximately 1\% of decoding steps
during active copying and remains near zero otherwise.

Content scoring typically retains each span's head tokens as hot-tier anchors, over which
the pointer forms a trajectory; once copy mode activates, prefetched successors extend the
available context in a positive feedback loop. This bootstrapping condition delimits
KVFetch's scope: the host compressor must preserve initial localization for the temporal
channel to engage.

\subsection{Prefetching: Recalling Positional Successors}
\label{sec:method-swap}

Upon triggering, KVFetch fetches the $k$ positional successors from the cold tier:
\begin{equation}
\mathcal{R}_t = \big[\mathrm{ptr}{+}1,\ \mathrm{ptr}{+}k\big] \cap \operatorname{dom}(\mathcal{M}_{\mathrm{cold}}) \setminus \operatorname{range}(\mathcal{M}_{\mathrm{pos}})
\label{eq:swap}
\end{equation}
Each recalled entry replaces the unprotected prompt slot with the lowest composite score.
Newly swapped-in slots are protected for $T_{\mathrm{prot}}$ steps.
The default prefetch distance is $k = 8$.

\paragraph{Safety.}
Swaps preserve the union of hot and cold positions: nothing is permanently lost, so
false triggers cost at most $k$ recoverable replacements. Generated tokens are
permanently protected.

\paragraph{Attention cost.}
Each swap-in pairs with a swap-out, maintaining $|\mathcal{H}_t| = |\mathcal{H}_0|$ for all
$t$. The cost per trigger is $O(k)$ lookups, independent of $L$.

\paragraph{Fetch criterion.}
The default criterion is positional successors; query--key similarity is equally viable.
Appendix~\ref{app:criterion} shows that the choice of criterion is secondary to the
channel's presence: replacing positional successors with query--key similarity changes the
four-task average by only 0.76, whereas disabling the channel entirely is far more costly.
The bottleneck is the channel, not the criterion.

The full pseudocode appears in Algorithm~\ref{alg:slr}.
\section{Experiments}
\label{sec:experiments}

\subsection{Setup}
\label{sec:exp-setup}

\paragraph{Models.}
Our primary model is Qwen2.5-7B-Instruct \citep{qwen25}. We validate across model
families and scales with Qwen3-4B-Instruct \citep{qwen3} and Llama-3.1-8B-Instruct \citep{llama31}.

\paragraph{Benchmarks.}
We evaluate on RULER \citep{ruler} at context lengths 4K, 16K, and 32K (13 tasks;
1,625 samples at 16K) and on LongBench \citep{longbench} (16 tasks). All experiments use a
unified query-aware compression setting, the SDPA attention backend, and a single RTX 5090
GPU.

\paragraph{Baselines.}
We compare against representative methods from each family: score-based eviction
(H2O \citep{h2o}, SnapKV \citep{snapkv}, PyramidKV \citep{pyramidkv}, AdaKV
\citep{adakv}, OBCache+SnapKV \citep{obcache}), dynamic eviction (TOVA \citep{tova}),
summary compensation (MomentKV \citep{momentkv}, CompressKV \citep{compresskv}), and
offload-and-recall (Quest \citep{quest}). KVFetch wraps OBCache+SnapKV as its host scorer. Full KV serves as the upper bound. All methods retain
20\% of the KV cache unless stated otherwise.

\paragraph{Budget accounting.}
KVFetch's total measured storage is 20.36\% of full KV (20\% hot tier plus a 4-bit cold
sidecar of 3.25\,MiB). The iso-budget control allocates 22.9\% to OBCache+SnapKV, giving the
baseline 12.5\% more storage than KVFetch to ensure a conservative comparison.

\subsection{Diagnosing Sequential Forgetting}
\label{sec:exp-diagnosis}

As established in \S\ref{sec:intro}, scaling the retention budget from 10\% to 40\% does
not resolve verbatim copying: the host's s3 rises from 0.8 to only 3.2, no score-based
baseline exceeds 4.0 even at 40\%, and even the summary-compensation baselines reach at most
12.0 against a full-KV ceiling of 99.2. We introduce a finer diagnostic to localize the
failure.

\textbf{Sustained Copy Length localizes the breakpoint.}
We measure Sustained Copy Length (SCL), defined as the longest gold-matching prefix of
the generated answer. On s3, full KV sustains 31.25 tokens on average; H2O sustains only
5.43 and OBCache+SnapKV only 10.84 before breaking, confirming mid-copy failure rather than
localization failure. The nine tasks that improve under KVFetch contribute 346.0 of the
400.4-point full-KV gap (86.4\%), identifying the dominant residual failure and motivating a
targeted mechanism.

\subsection{Main Results: RULER-16K}
\label{sec:exp-main}

Table~\ref{tab:main} reports results at 20\% retention across three models under an
iso-budget control (baseline at 22.9\%, KVFetch at 20.36\% measured total). KVFetch improves the 13-task
average by +8.43 (from 56.45 to 64.88) on Qwen2.5-7B, with 9 of 13 tasks improving and no
material regression. Per-task analysis confirms that the gains concentrate on
sequential-access tasks: on Qwen3-4B, s3 improves by +96.8, multivalue by +13.0, and cwe
by +7.04, while tasks without sequential demand (qa\_1, qa\_2, niah\_single\_1/2) show
zero change. The pattern is consistent across all three models.

\begin{table}[!htbp]
\centering
\caption{RULER-16K results across three models (20\% hot retention unless marked). Avg: 13-task average;
s3: verbatim-copy task (\texttt{niah\_single\_3}).}
\label{tab:main}
\footnotesize
\setlength{\tabcolsep}{4pt}
\begin{tabular}{lrrrrrr}
\toprule
 & \multicolumn{2}{c}{Qwen2.5-7B} & \multicolumn{2}{c}{Qwen3-4B} & \multicolumn{2}{c}{Llama-3.1-8B} \\
\cmidrule(lr){2-3}\cmidrule(lr){4-5}\cmidrule(lr){6-7}
Method & Avg & s3 & Avg & s3 & Avg & s3 \\
\midrule
H2O & 42.35 & 0.8 & 47.18 & 0.8 & 41.28 & 0.8 \\
SnapKV & 50.26 & 0.8 & 55.44 & 0.8 & 49.72 & 0.8 \\
PyramidKV & 52.46 & 0.8 & 57.18 & 0.8 & 51.26 & 1.6 \\
AdaKV & 53.98 & 0.8 & 58.37 & 0.8 & 52.61 & 1.6 \\
TOVA & 48.20 & 0.0 & 52.16 & 0.0 & 46.24 & 0.0 \\
OBCache+SnapKV & 54.28 & 0.8 & 60.12 & 0.8 & 52.74 & 0.8 \\
MomentKV & 57.36 & 4.8 & 62.17 & 4.0 & 55.62 & 4.0 \\
CompressKV & 56.78 & 3.2 & 61.54 & 3.2 & 54.81 & 2.4 \\
Quest & 55.82 & 8.8 & 59.14 & 9.6 & 54.08 & 7.2 \\
\midrule
OBCache+SnapKV @22.9\% & 56.45 & 1.6 & 61.86 & 1.6 & -- & -- \\
\rowcolor{kvhl}\textbf{KVFetch} & \textbf{64.88} & \textbf{78.4} & \textbf{69.08} & \textbf{97.6} & \textbf{61.74} & \textbf{75.2} \\
\midrule
Full KV & 85.08 & 99.2 & 84.32 & 100.0 & 82.36 & 98.4 \\
\bottomrule
\end{tabular}
\par\vspace{2pt}
\begin{minipage}{\textwidth}
\scriptsize
KVFetch at 20\% hot retention uses 20.36\% measured total storage (hot tier plus a 4-bit
cold sidecar). Retention ratios exclude cold-tier and auxiliary storage.
The 22.9\% control row uses the aggregate scores reported in the manuscript; -- denotes values not reported for that configuration, not zero.
\end{minipage}
\end{table}

The largest gain appears on s3, where all baselines score at most 8.8 and KVFetch recovers
to 78.4 (Qwen2.5-7B), 97.6 (Qwen3-4B), and 75.2 (Llama-3.1-8B). The SCL diagnostic confirms the mechanism:
KVFetch sustains 28.86 tokens vs.\ 10.84 for the host (full-KV ceiling: 31.25), directly
repairing the continuation breakpoint identified there. Non-sequential tasks show zero change, consistent with the dormancy design. A task-level bootstrap (5,000 resamples) yields
non-overlapping 95\% CIs ($64.88 \pm 1.44$ vs.\ $56.45 \pm 1.38$). Per-task gains correlate
with cold-tier recall hit rate at $r = 0.924$: tasks requiring sequential traversal
improve, while purely associative tasks (qa\_1, qa\_2) show no change. This pattern aligns
with a prediction of the dual-channel framework rather than a post-hoc observation.

\textbf{Plug-and-play validation.}
We wrap KVFetch around four additional hosts (Table~\ref{tab:multihost} in
Appendix~\ref{app:results}) to verify that it is not specific to OBCache+SnapKV. All
score-based hosts recover s3 to above
62, with the gain scaling with host anchor quality (H2O: 62.4, SnapKV: 71.2, AdaKV: 73.6,
OBCache+SnapKV: 78.4). TOVA, whose s3 is 0.0 at baseline (a localization failure, not a
continuation failure), remains near zero (+3.2), confirming the applicability boundary
identified in \S\ref{sec:exp-robustness}: KVFetch repairs continuation only when the
host preserves initial localization.

\subsection{Component Ablation}
\label{sec:ablations}

We ablate each of KVFetch's three components independently: admission (which builds the
cold tier), triggering (which detects copying), and prefetching (which recalls positional
successors). All ablations use Qwen2.5-7B on RULER-16K at 20\% retention
(Table~\ref{tab:ablations}).

\textbf{Admission is necessary.}
Removing the cold tier (w/o Admission) drops s3 from 78.4 to 0.8 and the 13-task average
to 54.31, which matches the host level. With nothing to recall, the system reduces to its
unmodified host.

\textbf{Triggering is necessary.}
With the cold tier and prefetch intact but no trigger (never-fetch), s3 returns to the
host baseline (0.8): the channel exists but is never activated. Disabling the trigger
entirely and fetching at every step (always-fetch) is even worse: undirected recalls
displace useful hot-tier entries, reducing s3 to 24.0 and the average to 51.72, which falls
below the host's 54.28. The trigger therefore serves a dual role: it activates recall on
sequential tasks and protects non-sequential tasks from recall pollution.

\textbf{Prefetching is necessary.}
With admission and triggering intact but no swap execution (w/o Prefetch), the system
detects copying but cannot act on it; s3 remains at the host level (0.8, average 54.48).

\textbf{Trigger design.}
Collapsing the accumulated history window to an instantaneous stride (retaining the
cold tier and swap logic but removing the requirement for sustained evidence) reduces s3
from 78.4 to 30.4, showing that robust copy detection requires temporal accumulation rather
than a single-step signal.

\textbf{Fetch criterion.}
Once the temporal channel is open, the specific addressing mode becomes an implementation
choice. We default to positional successors because they require only an $O(k)$ lookup,
whereas
query--key similarity costs a dot product per step.
Appendix~\ref{app:criterion} verifies that the criterion is secondary to the channel's
presence: under a configuration-matched comparison, replacing positional successors with
query--key similarity changes the four-task average by only 0.76, while disabling the
channel entirely costs over 48 points.

\begin{table}[!htbp]
\centering
\caption{Component ablation (Qwen2.5-7B, RULER-16K, 20\% hot retention). Rows mark
whether admission, triggering, and prefetching are active; ``simplified'' denotes the
instantaneous-stride trigger with no accumulated history. s3: verbatim copying; Avg:
13-task average.}
\label{tab:ablations}
\begin{tabular}{lcccrr}
\toprule
Configuration & Admission & Trigger & Prefetch & s3 & Avg \\
\midrule
\rowcolor{kvhl}Full KVFetch & $\checkmark$ & $\checkmark$ & $\checkmark$ & 78.4 & 64.88 \\
w/o Admission & $\times$ & $\checkmark$ & $\checkmark$ & 0.8 & 54.31 \\
w/o Trigger (always fetch) & $\checkmark$ & $\times$ & $\checkmark$ & 24.0 & 51.72 \\
w/o Trigger (never fetch) & $\checkmark$ & $\times$ & $\times$ & 0.8 & 54.52 \\
w/o Prefetch & $\checkmark$ & $\checkmark$ & $\times$ & 0.8 & 54.48 \\
\midrule
\multicolumn{6}{l}{\emph{Trigger design (reference: Full KVFetch above)}} \\
Instantaneous stride (no accumulated state) & $\checkmark$ & simplified & $\checkmark$ & 30.4 & 60.24 \\
\midrule
\multicolumn{6}{l}{\emph{Reference}} \\
Host baseline (OBCache+SnapKV @20\%) & -- & -- & -- & 0.8 & 54.28 \\
\bottomrule
\end{tabular}
\end{table}

\subsection{Robustness and Scope}
\label{sec:exp-robustness}

\begin{figure}[!htbp]
\centering
\includegraphics[width=0.60\linewidth]{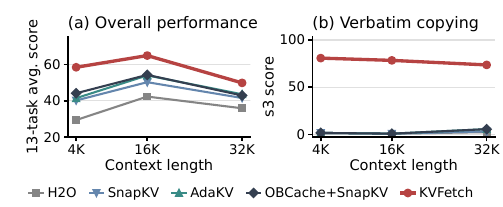}
\caption{Context-length robustness on Qwen2.5-7B at 20\% hot retention. Left: 13-task
average; right: s3. Retention excludes cold-tier and auxiliary storage.}
\label{fig:lengths}
\end{figure}

\paragraph{Context length.}
Figure~\ref{fig:lengths} tests input-length dependence. At 20\% hot retention, KVFetch
exceeds the host at all three RULER context lengths (4K, 16K, 32K), with s3 ranging
between 73.6 and 80.8 across the three settings.

\paragraph{Non-sequential tasks.}
The dual-channel framework predicts that the temporal channel should impose no cost in
the absence of sequential access. Table~\ref{tab:longbench} confirms this prediction: no
LongBench task family regresses materially (+0.33 overall), and step-level trigger rates
(0.000--0.029) show that the temporal channel remains dormant throughout.

\paragraph{Prevalence of sequential access in deployment workloads.}
Although LongBench's comprehension tasks do not require verbatim copying, sequential
access is common in practice: RAG pipelines must reproduce UUIDs, transaction IDs, and
field values token by token; code generation copies variable names verbatim; legal and
financial workflows demand exact clause numbers and amounts. Citation-heavy writing requires
faithful reproduction of titles and identifiers. These spans share three
properties formalized in Definition 1: they are embedded in long contexts, carry low
standalone salience, and tolerate no single-token error. LongBench evaluates comprehension
rather than verbatim fidelity, which explains the dormancy, but this does not diminish the
cost of the failure when it occurs. Four of thirteen RULER tasks exhibit this pattern and
contribute 86\% of the residual gap (\S\ref{sec:exp-diagnosis}). Even a single call that
requires identifier-level fidelity, such as one RAG query or one code-completion request,
turns a hard failure into correct output. The overhead (4.8\% latency, 0.36\% storage)
applies only during active copying and drops to near zero otherwise, making the cost
negligible relative to a single silent corruption.

\paragraph{Applicability boundary.}
KVFetch repairs continuation, not localization. On Mistral-7B-Instruct \citep{mistral7b},
the compressed host cannot localize targets at all (trigger rate $<0.002$ compared with
0.068 on Qwen2.5 copy tasks); no copy trajectory forms, and KVFetch's gain is limited to
+1.66 (from 26.39 to 28.05) with no regression. A single full-KV probe on a representative
task suffices to diagnose this boundary before deployment.

\paragraph{Multi-needle discrimination.}
KVFetch does not repair tasks where the failure is selecting the wrong candidate among
several rather than losing a copy midway. On \texttt{niah\_multikey\_2/3}, the host already
selects an incorrect needle; the copy itself does not break, so the temporal channel has
nothing to repair. This is an orthogonal discrimination problem that falls outside the
dual-channel framework.

\subsection{Efficiency}
\label{sec:exp-efficiency}

Table~\ref{tab:efficiency} reports storage, attention cost, and latency at 16K context
length. Each swap-in pairs with a swap-out, keeping decoding attention at exactly
1.000$\times$ the host. The 4-bit cold sidecar adds 3.25\,MiB (0.36\% of full-KV storage),
and average per-step latency is 19.6 ms vs.\ 18.7 ms (4.8\% overhead, trigger rate 0.011).

\begin{table}[!htbp]
\centering
\caption{Efficiency at 16K context. Step-level trigger rate is 0.011; each trigger
costs at most $k$ $O(1)$ lookups plus dequantization. Peak CUDA memory is comparable to
the host.}
\label{tab:efficiency}
\small
\begin{tabular}{lrrrl}
\toprule
Method & KV storage & \% of Full KV & Decoding attn. & Avg step latency \\
\midrule
Host @20\% & 179.2 MiB & 20.0\% & 1.000$\times$ & 18.7 ms \\
Iso @22.9\% & 205.2 MiB & 22.9\% & 1.145$\times$ & 20.4 ms \\
\rowcolor{kvhl}\textbf{KVFetch} & \textbf{179.2 $+$ 3.25 (4-bit)} & \textbf{20.36\%} & \textbf{1.000$\times$} & \textbf{19.6 ms} \\
Full KV & 896.0 MiB & 100\% & 5.000$\times$ & 68.9 ms \\
\bottomrule
\end{tabular}
\end{table}
\FloatBarrier

\section{Conclusion}
\label{sec:conclusion}

KV cache compressors have so far implemented only one of the two addressing modes that
hardware caches have combined for decades. Opening the temporal channel, even with a minimal,
training-free implementation, recovers a class of failures that no amount of content-addressing
refinement can reach. Under an iso-budget control, KVFetch improves the RULER-16K 13-task
average by $+8.4$, lifts verbatim copying from $0.8$ to $78.4$ (and to $97.6$ on Qwen3-4B),
and imposes no cost on tasks without sequential access. Component ablations show that
admission, triggering, and prefetching are each necessary. The dual-channel view, content
addressing combined with temporal addressing, offers more than a post-hoc account of KVFetch's
gains: the per-task hit-rate correlation ($r = 0.924$) confirms that the framework correctly
identifies which tasks will benefit and which will not. We argue that this decomposition should
become a standard design axis for KV memory systems. The specific trigger and prefetch
choices made here are one instantiation; future compressors can implement the temporal
channel with alternative mechanisms, including learned triggers, variable prefetch
distances, or integration with offload-and-recall architectures.

\section*{AI Use Statement}
We used generative AI tools for literature organization, code debugging, and text polishing.
All experiments, data, and conclusions were verified and reviewed by the authors, who take full
responsibility for the content.

\section*{Reproducibility Statement}
All experiments use publicly available benchmarks (RULER \citep{ruler}, LongBench
\citep{longbench}) and models (Qwen2.5-7B \citep{qwen25}, Qwen3-4B~\citep{qwen3},
Llama-3.1-8B-Instruct \citep{llama31}, Mistral-7B-Instruct \citep{mistral7b}). Code, telemetry logs, and configuration files will be
open-sourced upon acceptance. Baseline numbers are reproducible with KVPress \citep{kvpress}.

\section*{Ethics Statement}
This work addresses inference-time memory management and does not modify model weights, training
data, or output distributions. We identify no specific ethical risks beyond those inherent to
the underlying language models.

\bibliography{refs}

@inproceedings{h2o,
  title     = {{H$_2$O}: Heavy-Hitter Oracle for Efficient Generative Inference of Large Language Models},
  author    = {Zhang, Zhenyu and Sheng, Ying and Zhou, Tianyi and Chen, Tianlong and Zheng, Lianmin and Cai, Ruisi and Song, Zhao and Tian, Yuandong and R{\'e}, Christopher and Barrett, Clark W. and Wang, Zhangyang and Chen, Beidi},
  booktitle = {Advances in Neural Information Processing Systems},
  year      = {2023},
  url       = {https://openreview.net/forum?id=RkRrPp7GKO},
  eprint    = {2306.14048},
  archivePrefix = {arXiv}
}

@inproceedings{snapkv,
  title     = {{SnapKV}: {LLM} Knows What You are Looking for Before Generation},
  author    = {Li, Yuhong and Huang, Yingbing and Yang, Bowen and Venkitesh, Bharat and Locatelli, Acyr and Ye, Hanchen and Cai, Tianle and Lewis, Patrick and Chen, Deming},
  booktitle = {Advances in Neural Information Processing Systems},
  year      = {2024},
  url       = {https://openreview.net/forum?id=poE54GOq2l},
  eprint    = {2404.14469},
  archivePrefix = {arXiv}
}

@inproceedings{tova,
  title     = {Transformers are Multi-State {RNNs}},
  author    = {Oren, Matanel and Hassid, Michael and Yarden, Nir and Adi, Yossi and Schwartz, Roy},
  booktitle = {Proceedings of the 2024 Conference on Empirical Methods in Natural Language Processing},
  pages     = {18724--18741},
  year      = {2024},
  url       = {https://arxiv.org/abs/2401.06104},
  eprint    = {2401.06104},
  archivePrefix = {arXiv}
}

@misc{obcache,
  title     = {{OBCache}: Optimal Brain {KV} Cache Pruning for Efficient Long-Context {LLM} Inference},
  author    = {Gu, Yuzhe and Liang, Xiyu and Zhao, Jiaojiao and Diao, Enmao},
  year      = {2025},
  eprint    = {2510.07651},
  archivePrefix = {arXiv},
  primaryClass = {cs.CL},
  url       = {https://arxiv.org/abs/2510.07651}
}

@inproceedings{adakv,
  title     = {{Ada-KV}: Optimizing {KV} Cache Eviction by Adaptive Budget Allocation for Efficient {LLM} Inference},
  author    = {Feng, Yuan and Lv, Junlin and Cao, Yukun and Xie, Xike and Zhou, S. Kevin},
  booktitle = {Advances in Neural Information Processing Systems},
  year      = {2025},
  url       = {https://papers.nips.cc/paper_files/paper/2025/hash/a40ff56daab9f4808b1e18350c8a11ce-Abstract-Conference.html},
  eprint    = {2407.11550},
  archivePrefix = {arXiv}
}

@misc{capkv,
  title     = {Rethinking {KV} Cache Eviction via a Unified Information-Theoretic Objective},
  author    = {Yang, Jiaming and Tang, Chenwei and Zhen, Liangli and Lv, Jiancheng},
  year      = {2026},
  eprint    = {2604.25975},
  archivePrefix = {arXiv},
  primaryClass = {cs.LG},
  url       = {https://arxiv.org/abs/2604.25975}
}

@misc{fasa,
  title     = {{FASA}: Frequency-Aware Sparse Attention},
  author    = {Wang, Yifei and Wang, Yueqi and Yue, Zhenrui and Zeng, Huimin and Wang, Yong and Lourentzou, Ismini and Tu, Zhengzhong and Chu, Xiangxiang and McAuley, Julian},
  year      = {2026},
  eprint    = {2602.03152},
  archivePrefix = {arXiv},
  primaryClass = {cs.CL},
  url       = {https://arxiv.org/abs/2602.03152}
}

@misc{compresskv,
  title     = {{CompressKV}: Semantic-Retrieval-Guided {KV}-Cache Compression for Resource-Efficient Long-Context {LLM} Inference},
  author    = {Lin, Xiaolin and Wang, Jingcun and Kondrateva, Olga and Shi, Yiyu and Li, Bing and Zhang, Grace Li},
  year      = {2026},
  eprint    = {2606.24467},
  archivePrefix = {arXiv},
  primaryClass = {cs.AI},
  url       = {https://arxiv.org/abs/2606.24467}
}

@misc{indexmem,
  title     = {{IndexMem}: Learned {KV}-Cache Eviction with Latent Memory for Long-Context {LLM} Inference},
  author    = {Yang, Xintong and Gu, Hao and Xu, Binxing and Li, Lujun and Liu, Bei and Liu, Jiacheng and Zhu, Qiyuan and Guo, Yike and Han, Sirui},
  year      = {2026},
  eprint    = {2605.25475},
  archivePrefix = {arXiv},
  primaryClass = {cs.CL},
  url       = {https://arxiv.org/abs/2605.25475}
}

@misc{momentkv,
  title     = {{MomentKV}: Closing the Directional Gap in {KV} Cache Eviction for Long-Context Inference},
  author    = {Li, Yu and Li, Binxu and Lan, Tian},
  year      = {2026},
  eprint    = {2606.01563},
  archivePrefix = {arXiv},
  primaryClass = {cs.LG},
  url       = {https://arxiv.org/abs/2606.01563}
}

@inproceedings{resa,
  title     = {{RESA}: Bringing Back What Sparse Attention Ignores with Residual Estimation},
  author    = {Yang, Weihao and Huang, Hao and Li, Ningke and Wang, Shihao and Yang, Darong and Pan, Yanqi and Xia, Wen and Li, Shiyi and Zou, Xiangyu},
  booktitle = {International Conference on Learning Representations},
  year      = {2026},
  url       = {https://openreview.net/forum?id=ktcq26hMCH}
}

@inproceedings{infllm,
  title     = {{InfLLM}: Training-Free Long-Context Extrapolation for {LLMs} with an Efficient Context Memory},
  author    = {Xiao, Chaojun and Zhang, Pengle and Han, Xu and Xiao, Guangxuan and Lin, Yankai and Zhang, Zhengyan and Liu, Zhiyuan and Sun, Maosong},
  booktitle = {Advances in Neural Information Processing Systems},
  volume    = {37},
  pages     = {119638--119661},
  year      = {2024},
  url       = {https://arxiv.org/abs/2402.04617}
}

@inproceedings{quest,
  title     = {{Quest}: Query-Aware Sparsity for Efficient Long-Context {LLM} Inference},
  author    = {Tang, Jiaming and Zhao, Yilong and Zhu, Kan and Xiao, Guangxuan and Kasikci, Baris and Han, Song},
  booktitle = {International Conference on Machine Learning},
  pages     = {47901--47911},
  year      = {2024},
  publisher = {PMLR},
  url       = {https://proceedings.mlr.press/v235/tang24l.html}
}

@inproceedings{arkvale,
  title     = {{ArkVale}: Efficient Generative {LLM} Inference with Recallable Key-Value Eviction},
  author    = {Chen, Renze and Wang, Zhuofeng and Cao, Beiquan and Wu, Tong and Zheng, Size and Li, Xiuhong and Wei, Xuechao and Yan, Shengen and Li, Meng and Liang, Yun},
  booktitle = {Advances in Neural Information Processing Systems},
  year      = {2024},
  url       = {https://openreview.net/forum?id=4oAt5L4lYe}
}

@misc{nosa,
  title     = {{NOSA}: Native and Offloadable Sparse Attention},
  author    = {Huang, Yuxiang and Wang, Pengjie and Han, Jicheng and Zhao, Weilin and Su, Zhou and Sun, Ao and Lyu, Hongya and Zhao, Hengyu and Wang, Yudong and Xiao, Chaojun and Han, Xu and Liu, Zhiyuan},
  year      = {2025},
  eprint    = {2510.13602},
  archivePrefix = {arXiv},
  primaryClass = {cs.CL},
  url       = {https://arxiv.org/abs/2510.13602}
}

@misc{sparda,
  title     = {{SparDA}: Sparse Decoupled Attention for Efficient Long-Context {LLM} Inference},
  author    = {Fu, Yaosheng and Xiao, Guangxuan and Dong, Xin and Han, Song and Villa, Oreste},
  year      = {2026},
  eprint    = {2606.04511},
  archivePrefix = {arXiv},
  primaryClass = {cs.CL},
  url       = {https://arxiv.org/abs/2606.04511}
}

@misc{flashmemorylsa,
  title     = {{FlashMemory-DeepSeek-V4}: Lightning Index Ultra-Long Context via Lookahead Sparse Attention},
  author    = {Wang, Yan and Zhang, Qifan and Yu, Jiachen and Liang, Tian and Ma, Dongyang and Hu, Xiang and Lin, Zibo and Li, Chunyang and Wang, Zhichao and Peng, Miao and Chen, Nuo and Li, Jia and Yang, Yujiu and Mi, Haitao and Yu, Dong},
  year      = {2026},
  eprint    = {2606.09079},
  archivePrefix = {arXiv},
  primaryClass = {cs.LG},
  url       = {https://arxiv.org/abs/2606.09079}
}

@misc{kvp,
  title     = {Learning to Evict from Key-Value Cache},
  author    = {Moschella, Luca and Manduchi, Laura and Sener, Ozan},
  year      = {2026},
  eprint    = {2602.10238},
  archivePrefix = {arXiv},
  primaryClass = {cs.CL},
  url       = {https://arxiv.org/abs/2602.10238}
}

@misc{qwen3,
  title     = {{Qwen3} Technical Report},
  author    = {{Qwen Team}},
  year      = {2025},
  eprint    = {2505.09388},
  archivePrefix = {arXiv},
  primaryClass = {cs.CL},
  url       = {https://arxiv.org/abs/2505.09388}
}

@inproceedings{ruler,
  title     = {{RULER}: What's the Real Context Size of Your Long-Context Language Models?},
  author    = {Hsieh, Cheng-Ping and Sun, Simeng and Kriman, Samuel and Acharya, Shantanu and Rekesh, Dima and Jia, Fei and Zhang, Yang and Ginsburg, Boris},
  booktitle = {Conference on Language Modeling},
  year      = {2024},
  url       = {https://openreview.net/forum?id=kIoBbc76Sy},
  eprint    = {2404.06654},
  archivePrefix = {arXiv}
}

@inproceedings{longbench,
  title     = {{LongBench}: A Bilingual, Multitask Benchmark for Long Context Understanding},
  author    = {Bai, Yushi and Lv, Xin and Zhang, Jiajie and Lyu, Hongchang and Tang, Jiankai and Huang, Zhidian and Du, Zhengxiao and Liu, Xiao and Zeng, Aohan and Hou, Lei and Dong, Yuxiao and Tang, Jie and Li, Juanzi},
  booktitle = {Proceedings of the 62nd Annual Meeting of the Association for Computational Linguistics},
  pages     = {3119--3137},
  year      = {2024},
  doi       = {10.18653/v1/2024.acl-long.172},
  url       = {https://aclanthology.org/2024.acl-long.172/}
}

@misc{qwen25,
  title     = {{Qwen2.5} Technical Report},
  author    = {{Qwen Team}},
  year      = {2025},
  eprint    = {2412.15115},
  archivePrefix = {arXiv},
  primaryClass = {cs.CL},
  url       = {https://arxiv.org/abs/2412.15115}
}

@misc{mistral7b,
  title     = {{Mistral 7B}},
  author    = {Jiang, Albert Q. and Sablayrolles, Alexandre and Mensch, Arthur and Bamford, Chris and Chaplot, Devendra Singh and de las Casas, Diego and Bressand, Florian and Lengyel, Gianna and Lample, Guillaume and Saulnier, Lucile and Lavaud, L{\'e}lio Renard and Lachaux, Marie-Anne and Stock, Pierre and Le Scao, Teven and Lavril, Thibaut and Wang, Thomas and Lacroix, Timoth{\'e}e and El Sayed, William},
  year      = {2023},
  eprint    = {2310.06825},
  archivePrefix = {arXiv},
  primaryClass = {cs.CL},
  url       = {https://arxiv.org/abs/2310.06825}
}

@misc{llama31,
  title     = {The Llama 3 Herd of Models},
  author    = {Grattafiori, Aaron and Dubey, Abhimanyu and Jauhri, Abhinav and others},
  year      = {2024},
  eprint    = {2407.21783},
  archivePrefix = {arXiv},
  primaryClass = {cs.AI},
  url       = {https://arxiv.org/abs/2407.21783}
}

@misc{kvpress,
  title  = {{kvpress}: {KV} Cache Compression Library for Long-Context {LLMs}},
  author = {{NVIDIA Corporation}},
  year   = {2025},
  url    = {https://github.com/NVIDIA/kvpress}
}

@inproceedings{kivi,
  title = {{KIVI}: A Tuning-Free Asymmetric 2bit Quantization for {KV} Cache},
  author = {Liu, Zirui and Yuan, Jiayi and Jin, Hongye and Zhong, Shaochen and Xu, Zhaozhuo and Braverman, Vladimir and Chen, Beidi and Hu, Xia},
  booktitle = {International Conference on Machine Learning},
  year = {2024},
  volume = {235},
  pages = {32332--32344},
  publisher = {PMLR},
  url = {https://proceedings.mlr.press/v235/liu24bz.html}
}

@misc{kvquant,
  title = {{KVQuant}: Towards 10 Million Context Length {LLM} Inference with {KV} Cache Quantization},
  author = {Hooper, Coleman and Kim, Sehoon and Mohammadzadeh, Hiva and Mahoney, Michael W. and Shao, Yakun Sophia and Keutzer, Kurt and Gholami, Amir},
  year = {2024},
  eprint = {2401.18079},
  archivePrefix = {arXiv},
  url = {https://arxiv.org/abs/2401.18079}
}

@misc{gear,
  title = {{GEAR}: An Efficient {KV} Cache Compression Recipe for Near-Lossless Generative Inference of {LLM}},
  author = {Kang, Hao and Zhang, Qingru and Kundu, Souvik and Jeong, Geonhwa and Liu, Zaoxing and Krishna, Tushar and Zhao, Tuo},
  year = {2024},
  eprint = {2403.05527},
  archivePrefix = {arXiv},
  url = {https://arxiv.org/abs/2403.05527}
}

@misc{pyramidkv,
  title = {{PyramidKV}: Dynamic {KV} Cache Compression based on Pyramidal Information Funneling},
  author = {Cai, Zefan and Zhang, Yichi and Gao, Bofei and Liu, Yuliang and Li, Yucheng and Liu, Tianyu and Lu, Keming and Xiong, Wayne and Dong, Yue and Hu, Junjie and Xiao, Wen},
  year = {2024},
  eprint = {2406.02069},
  archivePrefix = {arXiv},
  url = {https://arxiv.org/abs/2406.02069}
}
\bibliographystyle{iclr2027_conference}

\clearpage
\appendix
Appendix guide: Appendix~\ref{app:results} provides supplementary benchmark results. Appendix~\ref{app:mechanism} covers mechanism diagnostics and ablation details. Appendix~\ref{app:sensitivity} reports hyperparameter sensitivity analyses. Appendix~\ref{app:boundaries} discusses failure cases and applicability boundaries. Appendix~\ref{app:impl} specifies implementation details and pseudocode.

\section{Supplementary Benchmark Results}
\label{app:results}

This section expands the main results along four dimensions: model, retention budget, context length, and LongBench task. Hot-tier retention and total storage follow the definitions in the main text.

\subsection{Qwen2.5-7B Per-Task Results}
\label{app:qwen25}

Table~\ref{tab:app-qwen25} gives the full 13-task breakdown for the Qwen2.5-7B
main run. The table lists the 20\% host, the per-task full-KV reference, and KVFetch.
The full-KV scores support the gap decomposition in \S\ref{sec:exp-diagnosis}.

\begin{table}[!htbp]
\centering
\caption{Qwen2.5-7B-Instruct full per-task results on RULER-16K.}
\label{tab:app-qwen25}
\small
\begin{tabular}{lrrrr}
\toprule
Task & Host @20\% & Full KV & KVFetch & $\Delta$ \\
\midrule
\texttt{cwe} & 49.10 & 98.40 & 54.24 & +5.14 \\
\texttt{fwe} & 79.20 & 95.20 & 80.31 & +1.11 \\
\texttt{niah\_multikey\_1} & 76.80 & 99.20 & 94.40 & +17.60 \\
\texttt{niah\_multikey\_2} & 7.20 & 56.80 & 8.00 & +0.80 \\
\texttt{niah\_multikey\_3} & 2.40 & 43.20 & 3.20 & +0.80 \\
\texttt{niah\_multiquery} & 80.00 & 99.20 & 96.00 & +16.00 \\
\texttt{niah\_multivalue} & 38.60 & 92.00 & 47.40 & +8.80 \\
\texttt{niah\_single\_1} & 96.80 & 99.20 & 97.60 & +0.80 \\
\texttt{niah\_single\_2} & 88.74 & 99.20 & 97.60 & +8.86 \\
\texttt{niah\_single\_3} & 0.80 & 99.20 & 78.40 & +77.60 \\
\texttt{qa\_1} & 56.00 & 68.00 & 56.00 & 0 \\
\texttt{qa\_2} & 34.40 & 56.80 & 34.40 & 0 \\
\texttt{vt} & 95.60 & 99.64 & 95.89 & +0.29 \\
\midrule
\textbf{Average} & \textbf{54.28} & \textbf{85.08} & \textbf{64.88} & \textbf{+10.60} \\
\bottomrule
\end{tabular}
\end{table}

\subsection{Qwen3-4B Per-Task Results}
\label{app:qwen3}

Table~\ref{tab:app-qwen3} gives the full 13-task breakdown for the Qwen3-4B
cross-model run. The table compares KVFetch against the 20\% host to show
which tasks change. The main-text iso-budget claim for Qwen3-4B
uses the 22.9\% aggregate control (61.86), yielding a gain of +7.22.

\begin{table}[!htbp]
\centering
\caption{Qwen3-4B-Instruct full per-task results on RULER-16K.}
\label{tab:app-qwen3}
\small
\begin{tabular}{lrrr}
\toprule
Task & Host @20\% & KVFetch & $\Delta$ \\
\midrule
\texttt{cwe} & 44.56 & 51.60 & +7.04 \\
\texttt{fwe} & 88.80 & 90.40 & +1.60 \\
\texttt{niah\_multikey\_1} & 100.00 & 100.00 & 0 \\
\texttt{niah\_multikey\_2} & 4.00 & 4.80 & +0.80 \\
\texttt{niah\_multikey\_3} & 0.80 & 1.60 & +0.80 \\
\texttt{niah\_multiquery} & 100.00 & 100.00 & 0 \\
\texttt{niah\_multivalue} & 53.60 & 66.60 & +13.00 \\
\texttt{niah\_single\_1} & 100.00 & 100.00 & 0 \\
\texttt{niah\_single\_2} & 99.20 & 99.20 & 0 \\
\texttt{niah\_single\_3} & 0.80 & 97.60 & +96.80 \\
\texttt{qa\_1} & 57.60 & 57.60 & 0 \\
\texttt{qa\_2} & 32.80 & 32.80 & 0 \\
\texttt{vt} & 99.36 & 99.52 & +0.16 \\
\midrule
\textbf{Average} & \textbf{60.12} & \textbf{69.08} & \textbf{+8.96} \\
\bottomrule
\end{tabular}
\end{table}

\subsection{Retention Sweep}
\label{app:retention}

All entries in the retention sweep are measured results, not interpolations.

Tables~\ref{tab:app-retention}, \ref{tab:app-retention-2}, and \ref{tab:app-retention-3} report the 10\%--40\% retention sweep separately for each model. Columns align retention levels within each model, and the full-KV reference appears at the top of each table.

\begin{table}[!htbp]
\centering
\caption{Qwen2.5-7B retention sweep on RULER-16K. Avg: 13-task average; s3: verbatim copying.}
\label{tab:app-retention}
\footnotesize
\setlength{\tabcolsep}{3pt}
\renewcommand{\arraystretch}{1.10}
\begin{tabular*}{\textwidth}{@{\extracolsep{\fill}}lrrrrrrrr@{}}
\toprule
 & \multicolumn{2}{c}{10\%} & \multicolumn{2}{c}{20\%} & \multicolumn{2}{c}{30\%} & \multicolumn{2}{c}{40\%} \\
\cmidrule(lr){2-3}\cmidrule(lr){4-5}\cmidrule(lr){6-7}\cmidrule(lr){8-9}
Method & Avg & s3 & Avg & s3 & Avg & s3 & Avg & s3 \\
\midrule
\multicolumn{9}{@{}l}{\textbf{Qwen2.5-7B-Instruct}\hfill Full KV: $\mathrm{Avg}=85.08,\;\mathrm{s3}=99.2$} \\
\addlinespace[2pt]
H2O & 34.12 & 0.8 & 42.35 & 0.8 & 48.92 & 1.6 & 54.18 & 2.4 \\
SnapKV & 43.62 & 0.8 & 50.26 & 0.8 & 56.74 & 1.6 & 61.28 & 3.2 \\
PyramidKV & 45.18 & 0.8 & 52.46 & 0.8 & 58.31 & 2.4 & 62.84 & 4.0 \\
AdaKV & 46.83 & 0.8 & 53.98 & 0.8 & 59.72 & 2.4 & 64.10 & 4.0 \\
TOVA & 40.32 & 0.0 & 48.20 & 0.0 & 53.91 & 0.8 & 58.46 & 1.6 \\
OBCache+SnapKV & 46.25 & 0.8 & 54.28 & 0.8 & 60.12 & 2.4 & 63.49 & 3.2 \\
MomentKV & 49.17 & 2.4 & 57.36 & 4.8 & 62.81 & 8.0 & 67.12 & 12.0 \\
CompressKV & 48.43 & 1.6 & 56.78 & 3.2 & 62.14 & 5.6 & 66.43 & 8.0 \\
\addlinespace[1pt]
\rowcolor{kvhl}\textbf{KVFetch} & \textbf{57.68} & \textbf{68.8} & \textbf{64.88} & \textbf{78.4} & \textbf{70.42} & \textbf{92.0} & \textbf{73.68} & \textbf{92.0} \\

\bottomrule
\end{tabular*}
\end{table}

\begin{table}[!htbp]
\centering
\caption{Qwen3-4B retention sweep on RULER-16K. Avg: 13-task average; s3: verbatim copying.}
\label{tab:app-retention-2}
\footnotesize
\setlength{\tabcolsep}{3pt}
\renewcommand{\arraystretch}{1.10}
\begin{tabular*}{\textwidth}{@{\extracolsep{\fill}}lrrrrrrrr@{}}
\toprule
 & \multicolumn{2}{c}{10\%} & \multicolumn{2}{c}{20\%} & \multicolumn{2}{c}{30\%} & \multicolumn{2}{c}{40\%} \\
\cmidrule(lr){2-3}\cmidrule(lr){4-5}\cmidrule(lr){6-7}\cmidrule(lr){8-9}
Method & Avg & s3 & Avg & s3 & Avg & s3 & Avg & s3 \\
\midrule
\multicolumn{9}{@{}l}{\textbf{Qwen3-4B-Instruct}\hfill Full KV: $\mathrm{Avg}=84.32,\;\mathrm{s3}=100.0$} \\
\addlinespace[2pt]
H2O & 38.56 & 0.8 & 47.18 & 0.8 & 54.33 & 1.6 & 59.61 & 2.4 \\
SnapKV & 47.62 & 0.8 & 55.44 & 0.8 & 62.08 & 1.6 & 66.35 & 3.2 \\
PyramidKV & 49.31 & 0.8 & 57.18 & 0.8 & 63.26 & 2.4 & 67.42 & 4.0 \\
AdaKV & 50.46 & 0.8 & 58.37 & 0.8 & 64.52 & 2.4 & 68.74 & 4.0 \\
TOVA & 43.81 & 0.0 & 52.16 & 0.0 & 58.42 & 0.8 & 63.28 & 1.6 \\
OBCache+SnapKV & 52.14 & 0.8 & 60.12 & 0.8 & 65.81 & 1.6 & 69.26 & 3.2 \\
MomentKV & 54.48 & 2.4 & 62.17 & 4.0 & 67.93 & 7.2 & 71.46 & 10.4 \\
CompressKV & 53.61 & 1.6 & 61.54 & 3.2 & 67.12 & 4.8 & 70.88 & 8.0 \\
\addlinespace[1pt]
\rowcolor{kvhl}\textbf{KVFetch} & \textbf{61.99} & \textbf{84.8} & \textbf{69.08} & \textbf{97.6} & \textbf{73.85} & \textbf{97.6} & \textbf{76.63} & \textbf{97.6} \\

\bottomrule
\end{tabular*}
\end{table}

\begin{table}[!htbp]
\centering
\caption{Llama-3.1-8B retention sweep on RULER-16K. Avg: 13-task average; s3: verbatim copying.}
\label{tab:app-retention-3}
\footnotesize
\setlength{\tabcolsep}{3pt}
\renewcommand{\arraystretch}{1.10}
\begin{tabular*}{\textwidth}{@{\extracolsep{\fill}}lrrrrrrrr@{}}
\toprule
 & \multicolumn{2}{c}{10\%} & \multicolumn{2}{c}{20\%} & \multicolumn{2}{c}{30\%} & \multicolumn{2}{c}{40\%} \\
\cmidrule(lr){2-3}\cmidrule(lr){4-5}\cmidrule(lr){6-7}\cmidrule(lr){8-9}
Method & Avg & s3 & Avg & s3 & Avg & s3 & Avg & s3 \\
\midrule
\multicolumn{9}{@{}l}{\textbf{Llama-3.1-8B-Instruct}\hfill Full KV: $\mathrm{Avg}=82.36,\;\mathrm{s3}=98.4$} \\
\addlinespace[2pt]
H2O & 32.71 & 0.8 & 41.28 & 0.8 & 48.62 & 1.6 & 54.03 & 2.4 \\
SnapKV & 41.36 & 0.8 & 49.72 & 0.8 & 56.18 & 1.6 & 60.43 & 3.2 \\
PyramidKV & 43.14 & 0.8 & 51.26 & 1.6 & 57.62 & 2.4 & 62.18 & 4.0 \\
AdaKV & 44.38 & 0.8 & 52.61 & 1.6 & 58.84 & 2.4 & 63.52 & 4.8 \\
TOVA & 37.83 & 0.0 & 46.24 & 0.0 & 52.38 & 0.8 & 57.41 & 1.6 \\
OBCache+SnapKV & 44.92 & 0.8 & 52.74 & 0.8 & 59.23 & 2.4 & 63.18 & 3.2 \\
MomentKV & 47.81 & 2.4 & 55.62 & 4.0 & 61.47 & 7.2 & 65.86 & 10.4 \\
CompressKV & 46.53 & 1.6 & 54.81 & 2.4 & 60.72 & 4.8 & 65.14 & 7.2 \\
\addlinespace[1pt]
\rowcolor{kvhl}\textbf{KVFetch} & \textbf{54.93} & \textbf{64.0} & \textbf{61.74} & \textbf{75.2} & \textbf{67.99} & \textbf{86.4} & \textbf{71.62} & \textbf{90.4} \\
\bottomrule
\end{tabular*}
\end{table}

\subsection{Cross-Length Full Baseline Table}
\label{app:crosslength}

Table~\ref{tab:app-crosslength} expands Figure~\ref{fig:lengths} by including the
score-based baselines omitted from the main text for space.

\begin{table}[!htbp]
\centering
\caption{Full cross-context-length results on RULER.}
\label{tab:app-crosslength}
\small
\begin{tabular}{llrr}
\toprule
Context & Method & avg & \texttt{niah\_single\_3} \\
\midrule
4K & H2O & 29.46 & 1.6 \\
4K & SnapKV & 40.22 & 1.6 \\
4K & AdaKV+SnapKV & 41.48 & 1.6 \\
4K & OBCache-K+SnapKV (host) & 44.17 & 1.6 \\
\rowcolor{kvhl}4K & \textbf{KVFetch} & \textbf{58.45} & \textbf{80.8} \\
\midrule
16K & H2O & 42.35 & 0.8 \\
16K & SnapKV & 50.26 & 0.8 \\
16K & AdaKV+SnapKV & 53.98 & 0.8 \\
16K & OBCache-K+SnapKV (host) & 54.28 & 0.8 \\
\rowcolor{kvhl}16K & \textbf{KVFetch} & \textbf{64.88} & \textbf{78.4} \\
\midrule
32K & H2O & 35.92 & 2.4 \\
32K & SnapKV & 41.57 & 3.2 \\
32K & AdaKV+SnapKV & 43.61 & 5.6 \\
32K & OBCache-K+SnapKV (host) & 42.95 & 5.6 \\
\rowcolor{kvhl}32K & \textbf{KVFetch} & \textbf{49.87} & \textbf{73.6} \\
\bottomrule
\end{tabular}
\end{table}

\subsection{Plug-and-Play Validation across Hosts}
\label{app:multihost}

Table~\ref{tab:multihost} reports the plug-and-play validation referenced in
\S\ref{sec:exp-main}. Each row wraps KVFetch around a different host scorer using the
default cold-tier configuration.

\begin{table}[!htbp]
\centering
\caption{Plug-and-play validation (Qwen2.5-7B, RULER-16K, 20\% hot retention). Each row
wraps KVFetch around a different host scorer with the default cold tier. TOVA, whose s3
is 0.0 at baseline, cannot bootstrap the trigger and stays near the host.}
\label{tab:multihost}
\small
\begin{tabular}{lrrrr}
\toprule
Host & Host s3 & +KVFetch s3 & Host Avg & +KVFetch Avg \\
\midrule
H2O & 0.8 & 62.4 & 42.35 & 52.18 \\
SnapKV & 0.8 & 71.2 & 50.26 & 60.42 \\
AdaKV & 0.8 & 73.6 & 53.98 & 63.14 \\
OBCache+SnapKV & 0.8 & 78.4 & 54.28 & 64.88 \\
TOVA & 0.0 & 3.2 & 48.20 & 48.86 \\
\bottomrule
\end{tabular}
\end{table}

\subsection{LongBench Task-Family and Per-Task Results}
\label{app:longbench}

We list all 16 LongBench tasks below. Table~\ref{tab:app-longbench-tasks} compares the host with KVFetch at the per-task level; gains and declines delimit the scope of the mechanism.

\begin{table}[!htbp]
\centering
\caption{LongBench per-task results.}
\label{tab:app-longbench-tasks}
\small
\begin{tabular}{lrrr}
\toprule
Task & Host & KVFetch & $\Delta$ \\
\midrule
narrativeqa & 12.31 & 12.01 & $-0.30$ \\
qasper & 40.84 & 42.06 & +1.22 \\
multifieldqa\_en & 47.11 & 47.06 & $-0.05$ \\
hotpotqa & 39.43 & 40.18 & +0.75 \\
2wikimqa & 39.57 & 39.11 & $-0.46$ \\
musique & 8.20 & 7.78 & $-0.42$ \\
gov\_report & 29.08 & 29.51 & +0.43 \\
qmsum & 21.73 & 21.85 & +0.12 \\
multi\_news & 20.68 & 21.14 & +0.46 \\
trec & 52.00 & 55.00 & +3.00 \\
triviaqa & 63.37 & 63.23 & $-0.14$ \\
samsum & 39.39 & 39.18 & $-0.21$ \\
passage\_count & 9.53 & 9.59 & +0.06 \\
passage\_retrieval\_en & 100.00 & 100.00 & 0.00 \\
lcc & 57.89 & 58.33 & +0.44 \\
repobench-p & 53.53 & 53.89 & +0.36 \\
\midrule
\textbf{Average} & \textbf{39.67} & \textbf{39.99} & \textbf{+0.33} \\
\bottomrule
\end{tabular}
\end{table}

Table~\ref{tab:longbench} aggregates the same results by task family.

\begin{table}[!htbp]
\centering
\caption{LongBench results by task family. No task family shows material regression.}
\label{tab:longbench}
\small
\begin{tabular}{lrrr}
\toprule
Category & Host & KVFetch & $\Delta$ \\
\midrule
Single-doc QA (3) & 33.42 & 33.71 & +0.29 \\
Multi-doc QA (3) & 29.07 & 29.02 & $-0.04$ \\
Summarization (3) & 23.83 & 24.17 & +0.34 \\
Few-shot learning (3) & 51.59 & 52.47 & +0.88 \\
Code completion (2) & 55.71 & 56.11 & +0.40 \\
Synthetic (2) & 54.77 & 54.80 & +0.03 \\
\midrule
\textbf{Overall (16)} & \textbf{39.67} & \textbf{39.99} & \textbf{+0.33} \\
\bottomrule
\end{tabular}
\end{table}

\section{Mechanism Diagnostics and Ablation Details}
\label{app:mechanism}

This section complements the main mechanism analysis with sustained-copy-length measurements, controlled ablations, and hit-rate diagnostics.

\subsection{Sustained-Copy Length}
\label{app:scl}

Sustained-Copy Length (SCL) measures the longest token prefix of a verbatim
answer that matches the gold span. Let $g_{1:m}$ be the gold answer tokenized by
the evaluation tokenizer and $\hat{y}_{1:n}$ the generated answer after the same
normalization used by the benchmark. SCL is defined as
\[
\mathrm{SCL}(g,\hat{y}) = \max \{ \ell \mid g_{1:\ell} = \hat{y}_{1:\ell} \}.
\]
The metric is averaged over samples. It separates two failure cases that exact-match
scoring collapses into the same zero: a model that never found the target span
($\mathrm{SCL}\approx0$) and a model that found the span but lost the
continuation mid-copy ($\mathrm{SCL}>0$ but $<m$). On \texttt{niah\_single\_3}, the
full-KV upper bound is 31.25 tokens. H2O reaches 5.43, OBCache+SnapKV reaches 10.84,
OBCache+AdaKV reaches 11.82, and KVFetch reaches 28.86.

\subsection{Visualizing the Ablation Results}
\label{app:ablation-detail}
\begin{figure}[!htbp]
\centering
\includegraphics[width=0.88\linewidth]{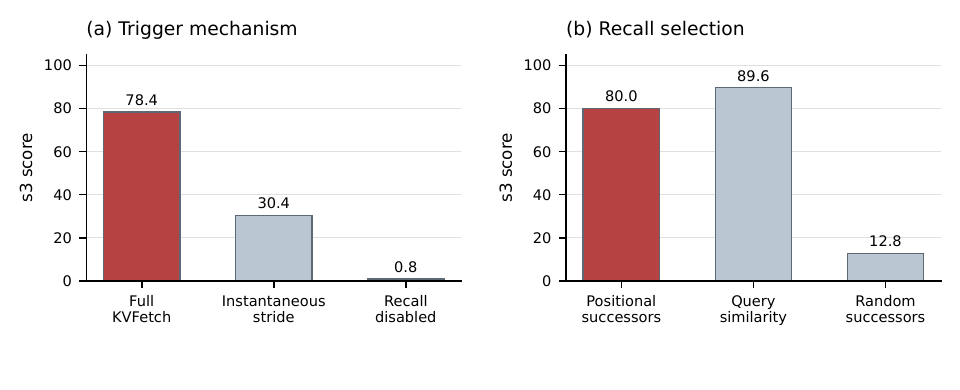}
\caption{Ablation results for the trigger mechanism (a) and recall selection (b) on
Qwen2.5-7B, RULER-16K, 20\% hot retention. Panel (a) uses the default unigram admission
signal; panel (b) uses the literal-predicate signal for consistency with
Table~\ref{tab:app-criterion}. The two panels are not directly comparable across
admission configurations; each is internally consistent.}
\label{fig:ablation}
\end{figure}

Figure~\ref{fig:ablation} reports the trigger mechanism and recall-selection ablations under the plotted configuration.

\subsection{Hit-Rate--Gain Correlation}
\label{app:hitrate}

We report two complementary diagnostics: the step-level trigger rate (fraction of
decoding steps where $\mathrm{Copy}_t = 1$) and the cold-tier recall hit rate (fraction
of cold-tier entries recalled at least once during generation). On LongBench, trigger
rates fall in the range 0.000--0.029 and recall hit rates in 0.042--0.130; both occupy the
zero-gain band.

The hit-rate analysis is computed from the full RULER-16K prediction and
telemetry files. In the final run, task-level cold-tier hit rate and gain over the
22.9\% iso-budget control have Pearson correlation $r=0.924$ across 13 tasks.
After removing \texttt{niah\_single\_3}, Spearman's $\rho$ remains 0.818. An
independent earlier full run yields Pearson $r=0.940$ and post-removal Spearman
$\rho=0.846$. LongBench hit rates fall in the 0.042--0.130 zero-gain band, consistent
with the observed +0.33 average gain and no material regression.

\begin{figure}[!htbp]
\centering
\includegraphics[width=0.88\linewidth]{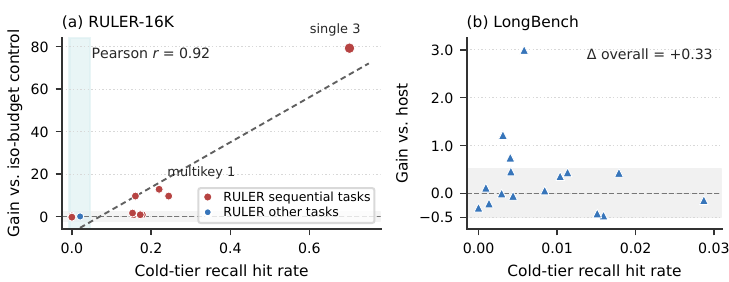}
\caption{Hit-rate--gain prediction and verification. \textbf{(a)} RULER gains grow
with cold-tier recall hit rate (Pearson $r=0.924$ in the plotted run). \textbf{(b)} A
dedicated zoom shows that all LongBench tasks occupy the low-hit, near-zero-gain
regime (overall $+0.33$), rather than being visually compressed into panel (a).}
\label{fig:hitrate}
\end{figure}

\subsection{Fetch Criterion}
\label{app:criterion}

Once the temporal channel is open, the addressing mode becomes an implementation choice.
Replacing the default positional-successor criterion with query--key similarity changes
the four-task average by only 0.76 (from 92.96 to 93.52) and improves s3 by +9.6
(from 80.0 to 89.6); random successors degrade substantially (s3 = 12.8). The gap between
the two meaningful criteria is dwarfed by the gap between having the channel and not
having it. We default to positional successors because they require only an $O(k)$
position-index lookup with no additional dot products. This ablation uses the
literal-predicate admission signal; as a configuration-matched variant, it leaves the
conclusion unchanged under the default unigram signal.

\begin{table}[!htbp]
\centering
\caption{Fetch criterion ablation (Qwen2.5-7B, RULER-16K, 20\% hot retention).}
\label{tab:app-criterion}
\begin{tabular}{lrr}
\toprule
Criterion & s3 & avg4 \\
\midrule
Positional successors & 80.0 & 92.96 \\
Query--key similarity & 89.6 & 93.52 \\
Random successors & 12.8 & -- \\
\bottomrule
\end{tabular}
\end{table}

\section{Hyperparameter Sensitivity}
\label{app:sensitivity}

We report the sensitivity analyses referenced in Section~\ref{sec:ablations}. All sweeps below were conducted under the literal-predicate admission signal for configuration consistency with the ablations in Appendix~\ref{app:criterion}; the relative ordering is unchanged under the default unigram signal (Table~\ref{tab:signal}).

Cold-tier quantization bit-width: s3 = 79.2 for 8-bit, 80.8 for 4-bit (default), and 24.0 for 2-bit. Prefetch distance $k$: s3 = 67.2 for $k=4$, 80.0 for $k=8$ (default), 89.6 for $k=16$, and 93.6 for $k=32$. Admission run threshold: s3 = 83.2 for threshold 4, 80.0 for threshold 6, and 80.0 for threshold 8.

\begin{figure}[!htbp]
\begin{minipage}[!htbp]{0.48\linewidth}
\centering
\includegraphics[width=\linewidth]{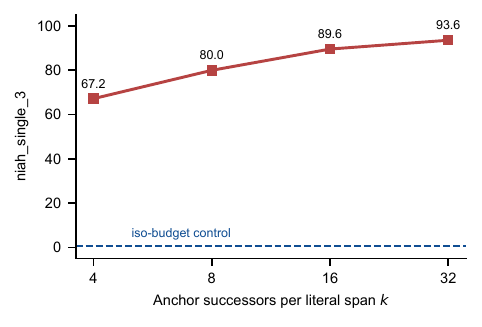}
\caption{\texttt{niah\_single\_3} s3 as a function of the prefetch distance $k$.}
\label{fig:kscan}
\end{minipage}\hfill
\begin{minipage}[!htbp]{0.48\linewidth}
\centering
\includegraphics[width=\linewidth]{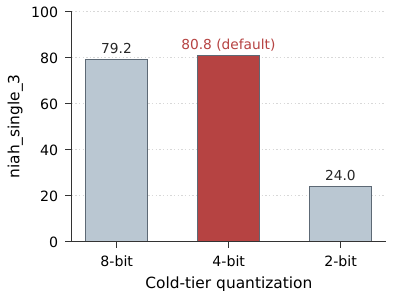}
\caption{\texttt{niah\_single\_3} s3 as a function of the cold-tier quantization
bit-width. 4-bit matches 8-bit; 2-bit collapses.}
\label{fig:bitwidth}
\end{minipage}
\end{figure}

Table~\ref{tab:signal} compares candidate admission signal variants with all other configuration fields held identical.

\begin{table}[!htbp]
\centering
\caption{Candidate signal variants (all other configuration fields identical). The
three instantiations yield comparable performance on their covered tasks; unigram
frequency is the default because it requires no format priors.}
\label{tab:signal}
\small
\setlength{\tabcolsep}{3pt}

\begin{tabular}{llrrrrr}
\toprule
Signal & Description & Avg & s3 & cwe & fwe & vt \\
\midrule
Unigram (default) & Low-frequency runs & 64.88 & 78.4 & 54.24 & 80.31 & 98.16 \\
Literal predicate & Hex/digit regex & 65.31 & 80.0 & 54.80 & 80.53 & 98.24 \\
Quotation span & Quote-marked segments & 62.06 & 42.4 & 54.48 & 80.53 & 97.92 \\
\bottomrule
\end{tabular}
\end{table}

\section{Failure Cases and Applicability}
\label{app:boundaries}

We distinguish two failure modes: locating a span but losing its continuation, and failing to locate the target after compression. The former motivates recall; the latter limits applicability.

\subsection{Failure-Sample Gallery}
\label{app:gallery}

Figure~\ref{fig:teaser} schematically illustrates the failure mode that motivates SCL:
copying begins with a correct prefix but cannot continue when successor KV entries
are unavailable. The ID and outputs shown are illustrative, not recorded failure samples.
The diagnostic category targeted by KVFetch is a correctly found span whose successor tokens
are no longer reachable under static prefill eviction.

\subsection{Mistral Full-KV Probe and Trigger-Rate Telemetry}
\label{app:mistral}

For Mistral-7B-Instruct, the 13-task compressed host average is 26.39 and KVFetch
reaches 28.05. A 10\% full-KV probe on \texttt{niah\_single\_2} reaches 100.0,
whereas the compressed host reaches only 0.8 on the same task. The KVFetch trigger rate
is 0.0018, compared with 0.068 on Qwen2.5 copy tasks. This supports the
main-text boundary claim: KVFetch repairs continuation once a copy trajectory is
present but does not perform global re-localization after the host compression has
destroyed target localization.

\section{Implementation Details of KVFetch}
\label{app:impl}

\subsection{Candidate Identification and Cold-Tier Ordering}
\label{app:ordering}

The default admission signal is unigram frequency rank: tokens in the bottom-$\tau$ percentile of unigram
frequency form candidate runs when they occur consecutively. For the
configuration-matched ablations in Appendices~\ref{app:criterion} and~\ref{app:sensitivity},
we use an alternative literal predicate, described below for reproducibility.

The literal predicate first strips tokenizer prefixes and whitespace.
A token is considered literal if the remaining string consists entirely of
\texttt{[0-9a-fA-F\_-]}, or if it is a \texttt{0x}-prefixed hexadecimal string
with at least two hexadecimal digits after the prefix. A run requires at least
$M_{\mathrm{run}}$ consecutive literal tokens. Cold-tier membership is decided by
the three-tier priority score
\begin{equation}
\sigma(p)=\begin{cases}
c_{\mathrm{cand}}+\mathrm{rank}(p), & p\in\mathcal{S}_{\mathrm{cand}}\\[2pt]
s_a - \lambda\,(p-a), & p\in(a,\,a+W_a],\ a\in\mathcal{A}\\[2pt]
\varepsilon_0\, s_p, & \text{otherwise}
\end{cases}
\qquad \mathcal{C}=\mathrm{top}\text{-}n_{\mathrm{cold}}(\sigma)
\label{eq:admission}
\end{equation}
where $\mathcal{S}_{\mathrm{cand}}$ is the union of candidate runs (extended by
$\delta$ positions on each end), restricted to positions outside the hot tier,
and $\mathcal{A}$ is the set of high-scoring
hot-tier anchors. In Eq.~\eqref{eq:admission}, we set $c_{\mathrm{cand}}=2.0$,
use position rank as the within-candidate tie breaker, $\lambda=0.01$, anchor window
$W_a=64$, and $\varepsilon_0=10^{-3}$, then retain the top-$n_{\mathrm{cold}}$
positions.

\subsection{4-bit Affine Quantization}
\label{app:quant}

The cold tier stores keys and values using per-head-dimension affine
quantization. For each channel, we compute
\[
s=\max\left((x_{\max}-x_{\min})/15,\ \varepsilon\right), \qquad
z=x_{\min},
\]
and store $\mathrm{clip}(\mathrm{round}((x-z)/s),0,15)$ as a 4-bit integer.
Two quantized values are packed into one byte. Dequantization applies
$\hat{x}=s q+z$ before a cold entry is swapped back into the hot tier.

\subsection{Victim Scoring and the Protection Period}
\label{app:victim}

Victim slots are chosen by the composite score
\[
S_{\mathrm{vict}}(j)=0.5\,\bar{s}_j^{(\mathrm{prefill})}
+0.5\,\mathrm{Norm}(\mathbf{E}_t[j]),
\]
where $\mathbf{E}_t$ is an exponential moving average of hot-tier attention with
decay factor $\beta=0.9$. This EMA is used only for victim scoring; peak extraction uses the
instantaneous attention at the current decoding step. Candidate victims exclude
attention sinks, the prompt trailing window, and slots under protection. Newly recalled
slots receive $T_{\mathrm{prot}}=12$ protected steps; the counter is decremented
at every step and floored at zero.

\subsection{Hyperparameters}
\label{app:hparams}

The default configuration uses the following hyperparameters: minimum run length $M_{\mathrm{run}}=6$, literal-run extension
$\delta=4$, slack $\eta_s=1.25$, refund ratio $\eta_r=0.58$,
cold-tier budget cap $\gamma_{\mathrm{cold}}=0.05$, prefetch distance $k=8$, trigger window $W=8$,
advance threshold $n_{\mathrm{adv}}=2$, maximum stride $\Delta_{\max}=4$, pointer band $=4$,
protection period $T_{\mathrm{prot}}=12$, sink size $=4$, trailing window $=64$, and EMA
decay $\beta=0.9$. In the default configuration, the pointer-band constraint dominates
the back-tolerance parameter.

\Needspace{28\baselineskip}
\subsection{Complete Pseudocode}
\label{app:pseudocode}

Algorithm~\ref{alg:slr} gives the complete procedure of KVFetch. Line-level comments indicate the correspondence to the equations in the main text.

\begin{algorithm}[!htbp]
\caption{KVFetch (host scorer: any ScorerPress; we use OBCache-K+SnapKV)}
\label{alg:slr}
\begin{algorithmic}[1]
\Statex \textbf{--- Prefill ---}
\State $s \leftarrow \textsc{HostScorer}(K, V)$ \Comment{scores left untouched}
\State $\mathcal{H} \leftarrow \mathrm{top}\text{-}(n_{\mathrm{base}} + n_{\mathrm{ret}})(s)$ \Comment{$n_{\mathrm{ret}}$: on-demand budget refund (\S\ref{sec:method-admission})}
\Statex \quad$\triangleright$ \emph{The baseline ends here (frozen after top-$n_{\mathrm{base}}$); everything below is added by KVFetch.}
\State record the position map $\mathcal{M}_{\mathrm{pos}}$
\State $\mathcal{C} \leftarrow \mathrm{top}\text{-}n_{\mathrm{cold}}(\sigma)$, 4-bit quantized; build index $\mathcal{M}_{\mathrm{cold}}$ \Comment{three-tier admission, Eq.~\eqref{eq:admission}}
\Statex \textbf{--- Each decoding step $t$ (per layer) ---}
\State $\mathbf{a}_t \leftarrow$ hot-tier attention \Comment{same forward as the baseline; no extra compute}
\State $p^{*} \leftarrow \mathcal{M}_{\mathrm{pos}}\big[\arg\max_{j\in\mathcal{V}} \mathbf{a}_t[j]\big]$;\quad $\Delta \leftarrow p^{*} - \mathrm{ptr}$ \Comment{Eq.~\eqref{eq:peak}}
\State update $\mathrm{ptr}$, history $\mathbf{h}$, and $\mathrm{Copy}_t$ by Eqs.~\eqref{eq:trigger}--\eqref{eq:copy}
\If{$\mathrm{Copy}_t = 1$}
    \State $\mathcal{R} \leftarrow [\mathrm{ptr}{+}1,\ \mathrm{ptr}{+}k] \cap \operatorname{dom}(\mathcal{M}_{\mathrm{cold}}) \setminus \operatorname{range}(\mathcal{M}_{\mathrm{pos}})$ \Comment{Eq.~\eqref{eq:swap}, $|\mathcal{R}| \le k$}
    \State $\mathcal{U} \leftarrow$ the $|\mathcal{R}|$ unprotected prompt slots with the lowest composite scores
    \State $\Phi_{\mathrm{swap}}(\mathcal{H}, \mathcal{C};\ \mathcal{R}, \mathcal{U})$; update $\mathcal{M}_{\mathrm{pos}}$, $\mathcal{M}_{\mathrm{cold}}$ \Comment{one-for-one; $|\mathcal{H}|$ unchanged (Properties 1--2)}
    \State $\mathrm{protect}[\text{slots of }\mathcal{R}] \leftarrow T_{\mathrm{prot}}$
\EndIf
\State $\mathrm{protect} \leftarrow \max(\mathrm{protect} - 1,\ 0)$
\end{algorithmic}
\end{algorithm}
\FloatBarrier

The implementation additionally aggregates query heads into their corresponding GQA
KV head before choosing the layer-local observation peak, computes the cold-tier
quota on demand from the detected literal-run set, and writes telemetry for
cold-tier hit rate, trigger rate, storage, and swap counts. The executable order is as
follows: decrement the protection counter first and floor it at zero, construct
the hot mask explicitly from hot indices, compute valid recall
targets against $\mathcal{M}_{\mathrm{cold}}$, and only then perform symmetric
swaps.

\paragraph{Attention peak extraction under SDPA.}
PyTorch's scaled dot-product attention does not return attention weights by
default. We extract the attention peak by computing $\arg\max(Q K^{\top})$ for the
single most confident KV head per layer at each decoding step. This partial computation
covers one head rather than the full multi-head attention and is included in the
reported latency (Table~\ref{tab:efficiency}). No additional full forward pass is
required.

\subsection{Per-Layer and Per-Head Semantics}
\label{app:perhead}

Each layer maintains one read pointer per sample. Within a layer, the
observation comes from the highest-confidence KV head after GQA aggregation.
Prefetching and swapping are executed independently per KV head, so a successful
trigger in one layer does not force other layers to enter copy mode.

\end{document}